\documentclass{article}
\usepackage{here,aclap,natbib,psfig}

\newtheorem{as}{Assumption}

\title{Automatic Discovery of Non-Compositional Compounds \\ in
Parallel Data \thanks{\hspace*{.05in} Many thanks to Mike Collins,
Jason Eisner, Mitch Marcus and two anonymous reviewers for their
feedback on earlier drafts of this paper.  This research was supported
by an equipment grant from Sun MicroSystems and by ARPA Contract
\#N66001-94C-6043.}
}

\author{I. Dan Melamed \\ Dept. of Computer and Information
Science \\ University of Pennsylvania \\ Philadelphia, PA, 19104,
U.S.A. \\ {\tt melamed@unagi.cis.upenn.edu} \\ {\tt http://www.cis.upenn.edu/\~{}melamed}}

\begin{document}

\maketitle

\begin{abstract}
Automatic segmentation of text into minimal content-bearing units is
an unsolved problem even for languages like English.  Spaces between
words offer an easy first approximation, but this approximation is not
good enough for machine translation (MT), where many word sequences
are not translated word-for-word.  This paper presents an efficient
automatic method for discovering sequences of words that are
translated as a unit.  The method proceeds by comparing pairs of
statistical translation models induced from parallel texts in two
languages.  It can discover hundreds of non-compositional compounds on
each iteration, and constructs longer compounds out of shorter ones.
Objective evaluation on a simple machine translation task has shown
the method's potential to improve the quality of MT output.
The method makes few assumptions about the data, so it can be applied
to parallel data other than parallel texts, such as word spellings and
pronunciations.
\end{abstract}

\section{Introduction}

The optimal way to analyze linguistic data into its primitive elements
is rarely obvious but often crucial.  Identifying phones and words in
speech has been a major focus of research.  Automatically finding
words in text, the problem addressed here, is largely unsolved for
languages such as Chinese and Thai, which are written without spaces
\citep[but see][]{fungwu,sproat}.  Spaces in texts of languages like
English offer an easy first approximation to minimal content-bearing
units.  However, this approximation mis-analyzes {\bf
non-compositional compounds (NCCs)} such as ``kick the bucket'' and
``hot dog.''  {\bf NCCs} are compound words whose meanings are a
matter of convention and cannot be synthesized from the meanings of
their space-delimited components.  Treating NCCs as multiple words
degrades the performance of machine translation (MT), information
retrieval, natural language generation, and most other NLP
applications.

NCCs are usually not translated literally to other languages.
Therefore, one way to discover NCCs is to induce and analyze a
translation model between two languages.  This paper is about an
information-theoretic approach to this kind of ontological discovery.
The method is based on the insight that treatment of NCCs as multiple
words reduces the predictive power of translation models.  Whether a
given sequence of words is an NCC can be determined by comparing the
predictive power of two translation models that differ on whether they
treat the word sequence as an NCC.  Searching a space of data models
in this manner has been proposed before, e.g.\ by \citet{ngramclus}
and \citet{langmod}, but their particular methods have been limited by
the computational expense of inducing data models and the typically
vast number of potential NCCs that need to be tested.  The method
presented here overcomes this limitation by making independence
assumptions that allow hundreds of NCCs to be discovered from each
pair of induced translation models.  It is further accelerated by
heuristics for gauging the {\em a priori} likelihood of validation for
each candidate NCC.

The predictive power of a translation model depends on what the model
is meant to predict.  This paper considers two different applications
of translation models, and their corresponding objective functions.
The different objective functions lead to different mathematical
formulations of predictive power, different heuristics for estimating
predictive power, and different classifications of word sequences with
respect to compositionality.  Monolingual properties of NCCs are not
considered by either objective function.  So, the method will not
detect phrases that are translated word-for-word despite
non-compositional semantics, such as the English metaphors ``ivory
tower'' and ``banana republic,'' which translate literally into
French.  On the other hand, the method will detect word sequences that
are often paraphrased in translation, but have perfectly compositional
meanings in the monolingual sense.  For example, ``tax system'' is
most often translated into French as ``r\'{e}gime fiscale.''  Each new
batch of validated NCCs raises the value of the objective function for
the given application, as demonstrated in Section~\ref{results}.  You
can skip ahead to Table~\ref{sample} for a random sample of the NCCs
that the method validated for use in a machine translation task.

The NCC detection method makes some assumptions about the properties
of statistical translation models, but no assumptions about the data
from which the models are constructed.  Therefore, the method is
applicable to parallel data other than parallel texts.  For example,
Section~\ref{results} applies the method to orthographic and phonetic
representations of English words to discover the NCCs of English
orthography.

\section{Translation Models}

A translation model can be constructed automatically from texts that
exist in two languages \mbox{({\bf bitexts})} \citep{ibm,transmod}.
The more accurate algorithms used for constructing translation models,
including the EM algorithm, alternate between two phases.  In the
first phase, the algorithm finds and counts the most likely links
between word tokens in the two halves of the bitext. {\bf Links}
connect words that are hypothesized to be mutual translations.  In the
second phase, the algorithm estimates translation probabilities by
dividing the link counts by the total number of links.  Let ${\cal S}$
and ${\cal T}$ represent the distributions of linked words in the
source and target\footnote{In the context of symmetric translation
models, the words ``source'' and ``target'' are merely labels.} texts.
A simple {\bf \mbox{translation} model} is just a joint probability
distribution $\Pr(s,t)$, which indicates the probability that a
randomly selected link in the bitext links $s \in {\cal S}$ with $t
\in {\cal T}$.\footnote{$s \in {\cal S}$ means that $\Pr_{\cal S}(s) >
0$.}  A {\bf directed \mbox{translation} model} can be derived in the
standard way: \linebreak $\Pr(t|s) = \Pr(s, t) / \Pr(s)$.

\section{Objective Functions}
\label{OF}
The decision whether a given sequence of words should count as an NCC
can be made automatically, if it can be expressed in terms of an
explicit objective function for the given application.  The first
application I will consider is statistical machine translation
involving a directed translation model and a target language model, of
the sort advocated by \citet{ibm}.  If only the translation model may be
varied, then the objective function for this application should be
based on how well the translation model predicts the distribution of
words in the target language.  In information theory, one such
objective function is called mutual information.  {\bf Mutual
information} measures how well one random variable predicts
another\footnote{See \citet{candt} for a good introduction to
information theory.}:
\begin{equation}
\label{midef}
I({\cal S}; {\cal T}) = \sum_{s \in {\cal S}} \sum_{t \in {\cal T}}
\Pr(s,t) \log \frac{\Pr(s,t)}{\Pr(s)\Pr(t)}
\end{equation}

When $\Pr(s,t)$ is a text translation model, mutual information
indicates how well the model can predict the distribution of words in
the target text given the distribution of words in the source text,
and vice versa.  This objective function may also be used for
optimizing cross-language information retrieval, where translational
distributions must be estimated either for queries or for documents
before queries and documents can be compared \citep{clir}.

\begin{figure}[htb]
\centering
\begin{tabular}{|c||r|l|}
\hline
Segment \# & English half & French half \\ \hline
1 & balance & \'{e}quilibre \\
2 & sheet & feuille \\
3 & balance sheet & bilan \\
\hline
\end{tabular}
\centerline{\psfig{figure=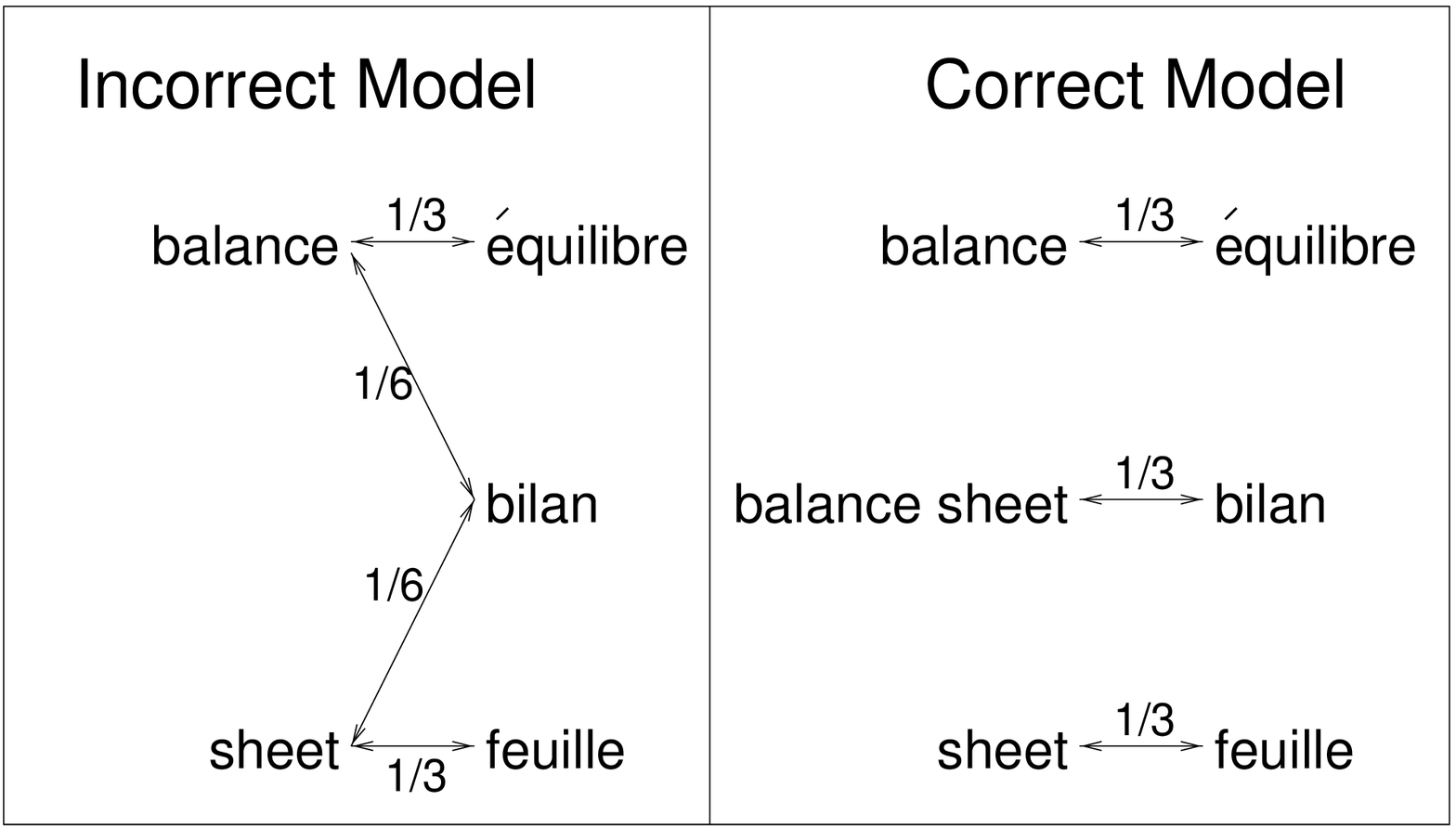,width=3in}}
\caption{{\em Two translation models that my be induced from the
trivial bitext at the top of the figure.  Translation models that know
about NCCs have higher mutual information than those that do
not.}\label{part}}
\end{figure}

Figure~\ref{part} shows a simple example of how
recognition of NCCs increases the mutual information of translation
models.  The English word ``balance'' is most often translated into
French as ``\'{e}quilibre'' and ``sheet'' usually becomes ``feuille.''
However, a ``balance sheet'' is a ``bilan.''  A translation model that
doesn't recognize ``balance sheet'' as an NCC would distribute the
translation probabilities of ``bilan'' over multiple English words, as
shown in the Incorrect Model.  The Incorrect Model is uncertain about
how ``bilan'' should be translated.  On the other hand, the Correct
Model, which recognizes ``balance sheet'' as an NCC is completely
certain about its translation.  As a result, the mutual information of
the Incorrect Model is $2 \cdot \frac{1}{3} \log
\frac{\frac{1}{3}}{\frac{1}{2} \cdot \frac{1}{3}} + 2 \cdot
\frac{1}{6} \log \frac{\frac{1}{6}}{\frac{1}{2} \cdot \frac{1}{3}} =
\frac{2}{3} \log 2$, whereas the mutual information of the Correct
Model is $\log 3$.

\section{Predictive Value Functions}
\label{pvf}
An explicit objective function immediately leads to a simple test of
whether a given sequence of words should be treated as an NCC: Induce
two translation models, a {\bf trial translation model} that involves
the candidate NCC and a {\bf base translation model} that does not.
If the value of the objective function is higher in the trial model
than in the base model, then the NCC is valid; otherwise it is not.
In theory, this test can be repeated for each sequence of words in the
text.  In practice, texts contain an enormous number of
word sequences \citep{ngramclus}, only a tiny fraction of which are
NCCs, and it takes considerable computational effort to induce
each translation model.  Therefore, it is necessary to test many NCCs
on each pair of translation models. 

Suppose we induce a trial translation model from texts $E$ and
$F$ involving a number of NCCs in the language ${\cal S}$ of $E$, and
compare it to a base translation model without any of those NCCs.  We
would like to keep the NCCs that caused a net increase in the
objective function $I$ and discard those that caused a net decrease.
We need some method of assigning credit for the difference in the
value of $I$ between the two models.  More precisely, we need a
function $i^{{\cal T}}(s)$ over the words $s \in {\cal S}$ such
that
\begin{equation}
\label{misum}
I({\cal S} ; {\cal T}) = \sum_{s \in {\cal S}} i^{{\cal T}}(s) .
\end{equation}
Fortunately, the objective function in Equations~\ref{midef} is
already a summation over source words.  So, its value can be
distributed as follows:
\begin{equation}
\label{i}
i^{{\cal T}}(s) = \sum_{t \in {\cal T}} \Pr(s,t) \log
\frac{\Pr(s,t)}{\Pr(s)\Pr(t)}
\end{equation}
The {\bf predictive value function} $i^{{\cal T}}(s)$ represents the
contribution of $s$ to the objective function of the whole translation
model.  I will write simply $i(s)$ when ${\cal T}$ is clear from the
context.

Comparison of predictive value functions across translation models
can only be done under
\begin{as}
\label{tester}
Treating the bigram \mbox{$<x,y>$} as an NCC will not affect the
predictive value function of any $s \in {\cal S}$ other than $x$, $y$,
and the NCC $xy$.
\end{as}
Let $i$ and $i'$ be the predictive value functions for source words in
the base translation model and in the trial translation model,
respectively.  Under Assumption~\ref{tester}, the net change in the
objective function effected by each candidate NCC $xy$ is
\begin{equation}
\label{Delta}
\Delta_{xy} = i'(x) + i'(y) + i'(xy) - i(x) - i(y) .
\end{equation}
If $\Delta_{xy} > 0$, then $xy$ is a valid NCC for the given
application.

Assumption~\ref{tester} would likely be false if either $x$ or $y$ was
a part of any candidate NCC other than $xy$.  Therefore, NCCs that are
tested at the same time must satisfy the {\bf mutual exclusion
condition}: No word $s \in {\cal S}$ may participate in more than one
candidate NCC at the same time.  Assumption~\ref{tester} may not be
completely safe even with this restriction, due to the imprecise
nature of translation model construction algorithms.

\section{Iteration}
\label{alg}

The mutual exclusion condition implies that multiple tests must be
performed to find the majority of NCCs in a given text.  Furthermore,
Equation~\ref{Delta} allows testing of only two-word NCCs.
Certainly, longer NCCs exist.  Given parallel texts $E$ and $F$, the
following algorithm runs multiple NCC tests and allows for recognition
of progressively longer NCCs:
\begin{enumerate}
\item Initialize the stop-list and the NCC list to be empty.
\item In $E$, find all occurrences of all NCCs on the NCC list, and
replace them with single ``fused'' tokens, which the translation model
construction algorithm will treat as single words.
\item Induce a base translation model between $E$ and~$F$.
\item For all adjacent bigrams $<x , y>$ in $E$ that are not on the
stop-list and whose frequency is at least $\phi$\footnote{The
threshold $\phi$ reduces errors due to noise in the data and in the
translation model.  It should be optimized empirically for each kind
of parallel data.  For parallel texts, I use $\phi = 2$.}, compute
$\hat{\Delta}_{xy}$, the estimate of $\Delta_{xy}$, using the
equations in Section~\ref{est}.
\item Make a list of candidate NCCs, containing all the bigrams for
which $\hat{\Delta}_{xy} > 0$, sorted by $\hat{\Delta}_{xy}$.
\item Remove from the list all candidates $xy$ where either $x$
or $y$ is part of another bigram higher in the list.  This step
implements the mutual exclusion condition described in Section~\ref{pvf}.
\item Copy $E$ to $E'$.  For each bigram $< x, y >$ remaining on the
candidate NCC list, fuse each instance of $<x,y>$ in $E'$ into a
single token $xy$.
\item Induce a trial translation model between $E'$ and~$F$.
\item Compute the actual $\Delta_{xy}$ values for all candidate NCCs,
using Equation~\ref{Delta}.
\item For each candidate NCC $xy$, if $\Delta_{xy} > 0$, then add $xy$
to the NCC list; otherwise add $xy$ to the stop-list.
\item Repeat from Step 2.
\end{enumerate}
The algorithm can also be run in ``two-sided'' mode, so that it looks
for NCCs in $E$ and in $F$ on alternate iterations.  This mode enables
the translation model to link NCCs in one language to NCCs in the
other.

In its simplest form, the algorithm only considers adjacent words as
candidate NCCs.  However, function words are translated very
inconsistently, and it is difficult to model their translational
distributions accurately.  To make discovery of NCCs involving
function words more likely, I consider content words that are
separated by one or two functions words to be adjacent.  Thus, NCCs
like \mbox{``blow ... whistle''} and \mbox{``icing ... cake''} may
contain gaps.

Fusing NCCs with gaps may fuse some words incorrectly, when the NCC is
a frozen expression.  For example, we would want to recognize that
``icing \ldots cake'' is an NCC when we see it in new text, but not if
it occurs in a sentence like ``Mary ate the icing off the cake.''  It
is necessary to determine whether the gap in a given NCC is fixed or
not.  Thus, the price for this flexibility provided by NCC gaps is
that, before Step 7, the algorithm fills gaps in proposed NCCs by
looking through the text.  Sometimes, NCCs have multiple possible gap
fillers, for example ``make up \{my,your,his,their\} mind.''  When the
gap filling procedure finds two or three possible fillers, the most
frequent filler is used, and the rest are ignored in the hope that
they will be discovered on the next iteration.  When there are more
than three possible fillers, the NCC retains the gap.  The token fuser
(in Steps 2 and~7) knows to shift all words in the NCC to the location
of the leftmost word.  E.g.\ an instance of the previous example in
the text might be fused as ``make\_up\_$<GAP>$\_mind his.''

In principle, the NCC discovery algorithm could iterate until
$\hat{\Delta}_{xy} < 0$ for all bigrams.  This would be a classic
case of over-fitting the model to the training data.  NCC discovery is
more useful if it is stopped at the point where the NCCs discovered so
far would maximize the application's objective function on new data.
A domain-independent method to find this point is to use held-out data
or, more generally, to cross-validate between different subsets of the
training data.  Alternatively, when the applications involves human
inspection, e.g.\ for bilingual lexicography, a suitable stopping point
can be found by manually inspecting validated NCCs.

\section{Credit Estimation}
\label{est}

\begin{figure*}[htb]
\begin{tabular*}{6.5in}{l}
\hline
\end{tabular*}
\begin{eqnarray}
\label{iprimexy}
i'(xy) & = & \sum_{t \in {\cal T}} \Pr(xy, t)
 \log \frac{ \Pr(xy, t)} {\Pr(xy) \Pr(t)} \\ 
\mbox{(by Eq.~\ref{prxy})}	& = & \sum_{t \in {\cal T}} [\Pr(x:\mbox{{\sc rc}} = y, t) +
\Pr(y:\mbox{{\sc lc}} = x, t) ] \log \frac{ [\Pr(x:\mbox{{\sc rc}} = y, t) + \Pr(y:\mbox{{\sc lc}} = x, t) ]
} {\Pr(y:\mbox{{\sc lc}} = x) \Pr(t)}  \nonumber \\ 
\mbox{(by Eq.~\ref{either})}	& = & \sum_{t \in {\cal T}}
 \Pr(x:\mbox{{\sc rc}} = y, t) \log \frac{\Pr(x:\mbox{{\sc rc}} = y,
 t)} {\Pr(x:\mbox{{\sc rc}} = y) \Pr(t)} \nonumber \\
\mbox{(by Eq.~\ref{or})}	& + & \sum_{t \in {\cal T}} \Pr(y:\mbox{{\sc lc}} = x, t)
\log \frac{\Pr(y:\mbox{{\sc lc}} = x, t)} {\Pr(y:\mbox{{\sc lc}} = x)
 \Pr(t)} \nonumber
\end{eqnarray}
\begin{tabular*}{6.5in}{l}
\hline
\end{tabular*}
\caption{{\em Estimation of  $i'(xy)$.  Note that, by definition,
$\Pr(x:RC = y) = \Pr(y:\mbox{{\sc lc}} = x) = \Pr(xy)$.} \label{fig:iprimexy}}
\end{figure*}
Sections~\ref{OF} and~\ref{pvf} describe how to carry out NCC validity
tests, but not how to choose which NCCs to test.  Making this choice
at random would make the discovery process too slow, because the vast
majority of word sequences are not valid NCCs.  The discovery process
can be greatly accelerated by testing only candidate NCCs for which
Equation~\ref{Delta} is likely to be positive.  This section presents
a way to guess whether $\Delta_{xy} > 0$ for a candidate NCC $xy$
\mbox{{\em before}} inducing a translation model that involves this
NCC.  To do so, it is necessary to estimate $i'(x)$, $i'(y)$, and
$i'(xy)$, using only the base translation model.

First, a bit of notation.  Let $\mbox{{\sc lc}}$ and $\mbox{{\sc rc}}$
denote word contexts to the left and to the right.  Let
\mbox{$(x:\mbox{{\sc rc}} = y)$} be the set of tokens of $x$ whose
right context is $y$, and vice versa for \mbox{$(y:\mbox{{\sc lc}} =
x)$}. Now, $i'(x)$ and $i'(y)$, can be estimated under
\begin{as}
\label{marginal}
When $x$ occurs {\em without} $y$ in its context, it will be linked to
the same target words by the trial translation model as by the base
translation model, and likewise for $y$ without $x$.
\end{as}
Assumption~\ref{marginal} says that
\begin{equation}
\label{fprimex}
i'(x) = i(x:\mbox{{\sc rc}} \neq y)
\end{equation}
\begin{equation}
\label{fprimey}
i'(y) = i(y:\mbox{{\sc lc}} \neq x)
\end{equation}

Estimating $i'(xy)$ is more difficult because it requires knowledge of
the entire translational distributions of both $x$ and $y$,
conditioned on all the contexts of $x$ and $y$.  Since we wish to
consider hundreds of candidate NCCs simultaneously, and contexts from
many megabytes of text, all this information would not fit on disk,
let alone in memory.  The best we can do is approximate with
lower-order distributions that are easier to compute.

The approximation begins with
\begin{as}
\label{as3}
If $xy$ is a valid NCC, then at most one of $x$ and $y$ will be
linked to a target word whenever $x$ and $y$ co-occur.
\end{as}
Assumption~\ref{as3} implies that for all $t \in {\cal T}$
\begin{equation}
\label{prxy}
\hspace*{-.09in}\Pr(xy, t) = \Pr(x:\mbox{{\sc rc}} = y, t) + \Pr(y:\mbox{{\sc lc}} = x, t) 
\end{equation}
The approximation continues with
\begin{as}
\label{eitheror}
If $xy$ is a valid NCC, then for all $t \in {\cal T}$, either $\Pr(x,
t) = 0$ or $\Pr(y, t) = 0$.
\end{as}
Assumption~\ref{eitheror} also implies that for all $t \in {\cal T}$,
either 
\begin{equation}
\label{either}
\Pr(x:\mbox{{\sc rc}} = y, t) = 0 
\end{equation}
or 
\begin{equation}
\label{or}
\Pr(y:\mbox{{\sc lc}} = x, t) = 0.
\end{equation}
Under Assumptions 3 and 4, we can estimate $i'(xy)$ as shown in
Figure~\ref{fig:iprimexy}.

The final form of Equation~\ref{iprimexy} (in Figure~\ref{fig:iprimexy})
allows us to partition all the terms in Equation~\ref{Delta} into two
sets, one for each of the components of the candidate NCC:
\begin{equation}
\label{Deltasum}
\hat{\Delta}_{xy} = \hat{\Delta}_{x \rightarrow y} + \hat{\Delta}_{x \leftarrow y}
\end{equation}
\newpage
where
\begin{eqnarray}
\label{rightDelta}
\hat{\Delta}_{x \rightarrow y} \hspace*{-.1in} & = & \hspace*{-.1in} -\mbox{} i(x) \\ \nonumber
\hspace*{-.1in} & + & \hspace*{-.1in}  \sum_{t \in {\cal T}} \Pr(x:\mbox{{\sc rc}} \neq y, t) \log
\frac{\Pr(x:\mbox{{\sc rc}} \neq y, t)}{\Pr(x, \mbox{{\sc rc}} \neq y)\Pr(t)} \\
\nonumber
\hspace*{-.1in} & + & \hspace*{-.1in} \sum_{t \in {\cal T}} \Pr(x:\mbox{{\sc rc}} = y, t)
\log \frac{\Pr(x:\mbox{{\sc rc}} = y, t)} {\Pr(x:\mbox{{\sc rc}} = y) \Pr(t)}
\end{eqnarray}
\begin{eqnarray}
\label{leftDelta}
\hat{\Delta}_{x \leftarrow y} \hspace*{-.1in} & = & \hspace*{-.1in}  -\mbox{} i(y) \\ \nonumber
\hspace*{-.1in} & + & \hspace*{-.1in}  \sum_{t \in {\cal T}} \Pr(y:\mbox{{\sc lc}} \neq x,t)
\log \frac{\Pr(y:\mbox{{\sc lc}} \neq x,t)}{\Pr(y, \mbox{{\sc lc}} \neq x)
\Pr(t)} \\ \nonumber
\hspace*{-.1in} & + & \hspace*{-.1in}  \sum_{t \in {\cal T}} \Pr(y:\mbox{{\sc lc}} = x, t)
\log \frac{\Pr(y:\mbox{{\sc lc}} = x, t)} {\Pr(y:\mbox{{\sc lc}} = x) \Pr(t)}
\end{eqnarray}
All the terms in Equation~\ref{rightDelta} depend only on the
probability distributions $\Pr(x, t)$, \mbox{ $\Pr(x:\mbox{{\sc rc}} =
y, t)$} and~\mbox{ $\Pr(x:\mbox{{\sc rc}} \neq y, t)$}. All the terms
in Equation~\ref{leftDelta} depend only on $\Pr(y, t)$, \mbox{
$\Pr(y:\mbox{{\sc lc}} = x, t)$} and~\mbox{ $\Pr(y:\mbox{{\sc lc}}
\neq x, t)$}.  These distributions can be computed efficiently by
memory-external sorting and streamed accumulation.  

\section{Bag-of-Words Translation}
\label{MT}
In bag-of-words translation, each word in the source text is simply
replaced with its most likely translation.  No target language model
is involved.  For this application, it is sufficient to predict only
the maximum likelihood translation of each source word.  The rest of
the translational distribution can be ignored.  Let $m^{{\cal T}}(s)$
be the most likely translation of each source word $s$, according to
the translation model:
\begin{equation}
\label{mdef}
m^{{\cal T}}(s) = \arg \max_{t \in T} \Pr(s, t)
\end{equation}
Again, I will write simply $m(s)$ when ${\cal T}$ is clear from the
context.  The objective function $V$ for this application follows by
analogy with the mutual information function $I$ in
Equation~\ref{midef}:
\begin{eqnarray}
\label{ofdef}
V({\cal S}; {\cal T}) \hspace*{-.1in} & = & \hspace*{-.1in} \sum_{s \in {\cal S}} \sum_{t \in {\cal
T}} \delta(t, m(s)) \Pr(s,t) \log \frac{\Pr(s,t)}{\Pr(s)\Pr(t)} \nonumber \\ 
\hspace*{-.1in} & = & \hspace*{-.1in} \sum_{s \in {\cal S}} \Pr(s,m(s)) \log
\frac{\Pr(s,m(s))}{\Pr(s)\Pr(m(s))}
\end{eqnarray}
The Kronecker $\delta$ function is equal to one when its arguments are
identical and zero otherwise.

The form of the objective function again permits easy distribution of
its value over the $s \in {\cal S}$:
\begin{equation}
\label{v}
v^{{\cal T}}(s) = \Pr(s,m(s)) \log \frac{\Pr(s,m(s))}{\Pr(s)\Pr(m(s))}
.
\end{equation}

The formula for estimating the net change in the objective function
due to each candidate NCC remains the same:
\begin{equation}
\label{DeltaV}
\Delta_{xy} = v'(x) + v'(y) + v'(xy) - v(x) - v(y) .
\end{equation}
It is easier to estimate the values of $v'$ using only the base
translation model, than to estimate the values of $i'$, since only the
most likely translations need to be considered, instead of entire
translational distributions.  $v'(x)$ and $v'(y)$ are again
estimated under Assumption~\ref{marginal}:
\begin{equation}
\label{vprimex}
v'(x) = v(x:\mbox{{\sc rc}} \neq y)
\end{equation}
\begin{equation}
\label{vprimey}
v'(y) = v(y:\mbox{{\sc lc}} \neq x)
\end{equation}
$v'(xy)$ can be estimated without making the strong
assumptions~\ref{as3} and~\ref{eitheror}.  Instead, I use the weaker
\begin{as}
\label{as5}
Let $t_x$ and $t_y$ be the most frequent translations of $x$ and $y$
in each other's presence, in the base translation model.
The most likely translation of $xy$ in the trial translation model
will be the more frequent of $t_x$ and $t_y$.
\end{as}
Assumption~\ref{as5} implies that
\begin{equation}
\label{vprimexy}
v'(xy) = max[ v(x:\mbox{{\sc rc}} = y) , v(y:\mbox{{\sc lc}} = x)] .
\end{equation}
This quantity can be computed exactly at a reasonable computational
expense.

\section{Experiments}
\label{results}

To demonstrate the method's applicability to data other than parallel
texts, and to illustrate some of its interesting properties, I
describe my last experiment first.  I applied the mutual information
objective function and its associated predictive value function to a
data set consisting of spellings and pronunciations of 17381 English
words.  Table~\ref{ortho} shows the NCCs of English spelling that the
algorithm discovered on the first 10 iterations.
\begin{table}[htb]
\centering
\begin{tabular}{|r|l|l|}
\hline
Iteration & Validated NCCs & Example \\ \hline
1 & er & father, her \\
& ng & hang \\
& ch  & chat, school\\
& ou  & court, could \\ \hline
2 & es & files \\
& au  & august \\ 
& gh  & laugh \\ \hline
3 & th  & this, thin \\
& ough & though, through \\ \hline
4 & (none)  & \\ \hline
5 & sh  & share \\ \hline
6 & io & tension \\ 
& ph  & graph \\ \hline
7 & tio  & nation \\
& ow  & know, how \\ 
& ck  & stack \\ \hline
8 & ea & near \\
& oo & book, tool \\
& ess & dress \\ \hline
9 & ia  & partial, facial \\ \hline
10 & (none) & \\
\hline
\end{tabular}
\caption{{\em The NCCs of English orthography discovered by the
algorithm.} \label{ortho}}
\end{table}  
The table reveals some interesting behavior of the algorithm.  The
NCCs ``er,'' ``ng'' and ``ow'' were validated because this data set
represents the sounds usually produced by these letter combinations with
one phoneme.  The NCC ``es'' most often appears in word-final
position, where the ``e'' is silent.  However, when ``es'' is not
word-final, the ``e'' is usually not silent, and the most frequent
following letter is ``s'', which is why the NCC ``ess'' was
validated.  NCCs like ``tio'' and ``ough'' are built up over multiple
iterations, sometimes out of pairs of previously discovered NCCs.

The other two experiments were carried out on transcripts of Canadian
parliamentary debates, known as the Hansards.  French and English
versions of these texts were aligned by sentence using the method of
\citet{align}.  Morphological variants in both languages were stemmed
to a canonical form.  Thirteen million words (in both languages
combined) were used for training and another two and a half million
were used for testing.  All translation models were induced using the
method of \citet{transmod}.  Six iterations of the NCC discovery
algorithm were run in ``two-sided'' mode, using the objective function
$I$, and five iterations were run using the objective function $V$.
Each iteration took approximately 78 hours on a 167MHz UltraSPARC
processor, running unoptimized Perl code.

\begin{table*}
\centering
\begin{tabular}{c|c|c||c|c|c}
\hline
Iteration & Bitext  & Vocabulary & Number of  & Number of & Validation \\
Number    & Side & 	Size 	& Proposed NCCs& Accepted NCCs	& Rate \\
\hline\hline
1	&	English &	29617 &	647 &		105 & 16\% \\
2	&	French & 	31664 &	618 &		121 &	20\% \\
3	&	English & 	29691 &	253 &		49 &	19\% \\
4	&	French & 	31768 &	245 &		41 &	17\% \\
5	&	English & 	29739 &	161 &		38 &	24\% \\
6	&	French & 	31809 &	205 &		33 &	16\% \\
\hline
\end{tabular}
\caption{{\em NCCs proposed and accepted, using the mutual information
objective function $I$.} \label{mitable}}
\end{table*}
\begin{table*}
\centering
\begin{tabular}{c|c|c||c|c|c}
\hline
Iteration & Bitext  & Vocabulary & Number of  & Number of & Validation \\
Number    & Side & 	Size & Proposed NCCs & Accepted NCCs & Rate \\
\hline\hline
1	&	English &	29617 &	776 &		758 & 98\% \\
2	&	French & 	31664 &	758 &		748 & 99\% \\
3	&	English & 	30333 &	399 &		388 & 97\% \\
4	&	French & 	32384 &	355 &		340 & 96\% \\
5	&	English & 	30711 &	300 &		286 & 95\% \\
\hline
\end{tabular}
\caption{{\em NCCs proposed and accepted, using the simpler
objective function $V$.} \label{pvtable}}
\end{table*}
Tables~\ref{mitable} and~\ref{pvtable} chart the NCC discovery
process.  The NCCs proposed for the $V$ objective function were much
more likely to be validated than those proposed for $I$, because the
predictive value function $v'$ is much easier to estimate {\em a
priori} than the predictive value function $i'$.  In 3 iterations on
the English side of the bitext, 192 NCCs were validated for $I$ and
1432 were validated for $V$.  Of the 1432 NCCs validated for $V$, 84
NCCs consisted of 3 words, 3 consisted of 4 words and 2 consisted of 5
words.  The French NCCs were longer on average, due to the frequent
``N de N'' construction for noun compounds.

\begin{figure}[htb]
\centerline{\psfig{figure=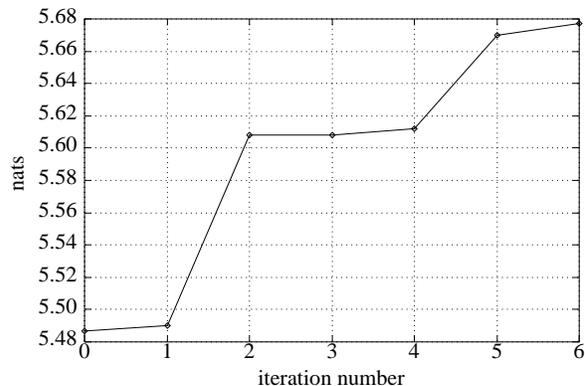,width=3in}}
\caption{{\em Mutual information of successive translation models
induced on held-out test data.  Nats are a measure of information like
bits, but based on the natural logarithm.  Translation models that
know about NCCs have higher mutual information than those that do
not.}\label{miinc}}
\end{figure}

The first experiment on the Hansards involved the mutual information
objective function $I$ and its associated predictive value function in
Equation~\ref{i}.  The first step in the experiment was the
construction of 5 new versions of the test data, in addition to the
original version.  Version $k$ of the test data was constructed by
fusing all NCCs validated up to iteration $k$ on the training data.
The second step was to induce a translation model from each version of
the test data.  There was no opportunity to measure the impact of NCC
recognition under the objective function $I$ on any real application,
but Figure~\ref{miinc} shows that the mutual information of successive
test translation models rose as desired.

\begin{figure}[htb]
\centerline{\psfig{figure=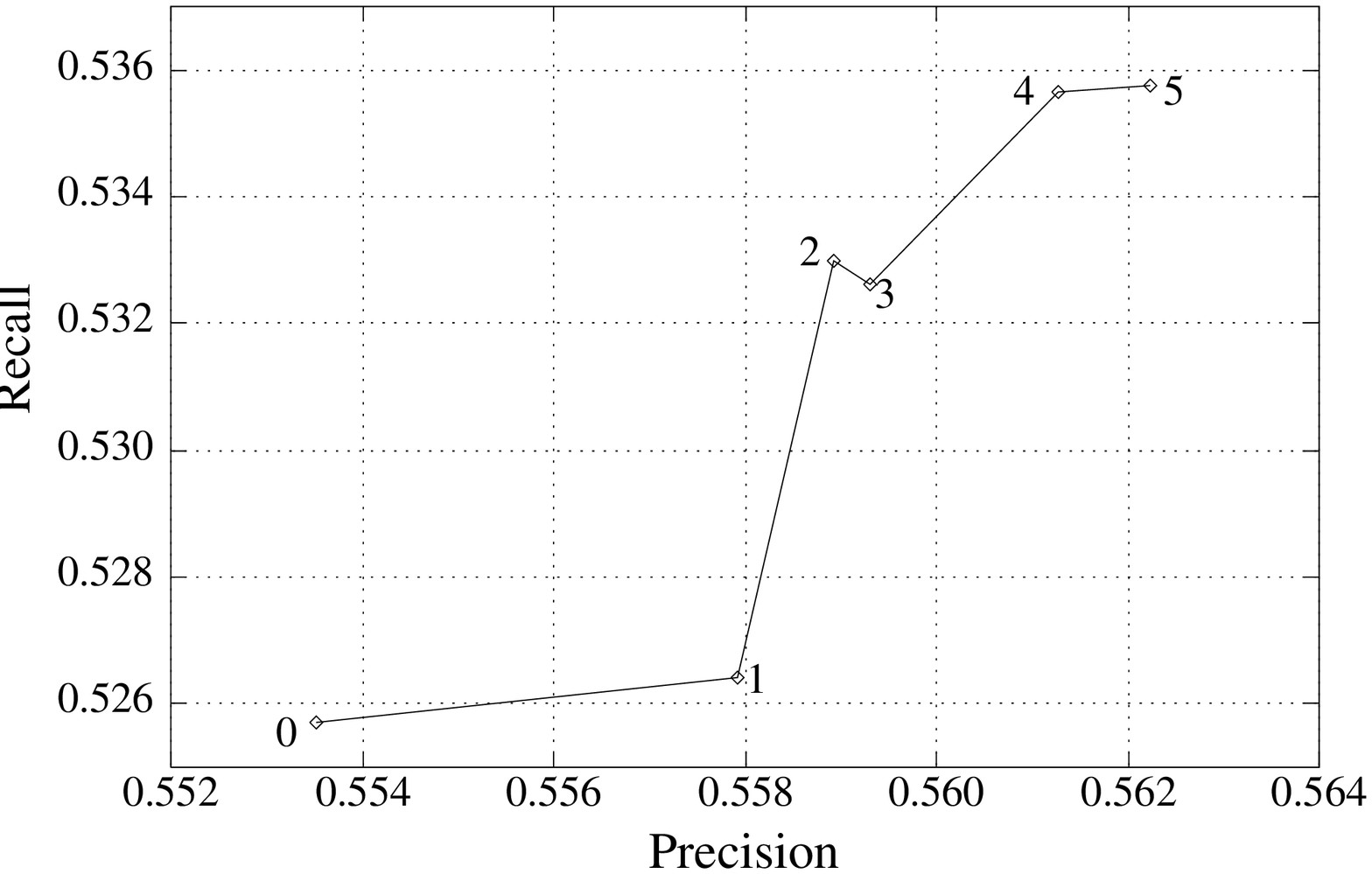,width=3in}}
\caption{{\em English $\rightarrow$ French BiBLE scores for 6
translation models.  Labels 0 to 5 indicate iteration number.}\label{biblenonulls}}
\end{figure}
\begin{figure}[htb]
\centerline{\psfig{figure=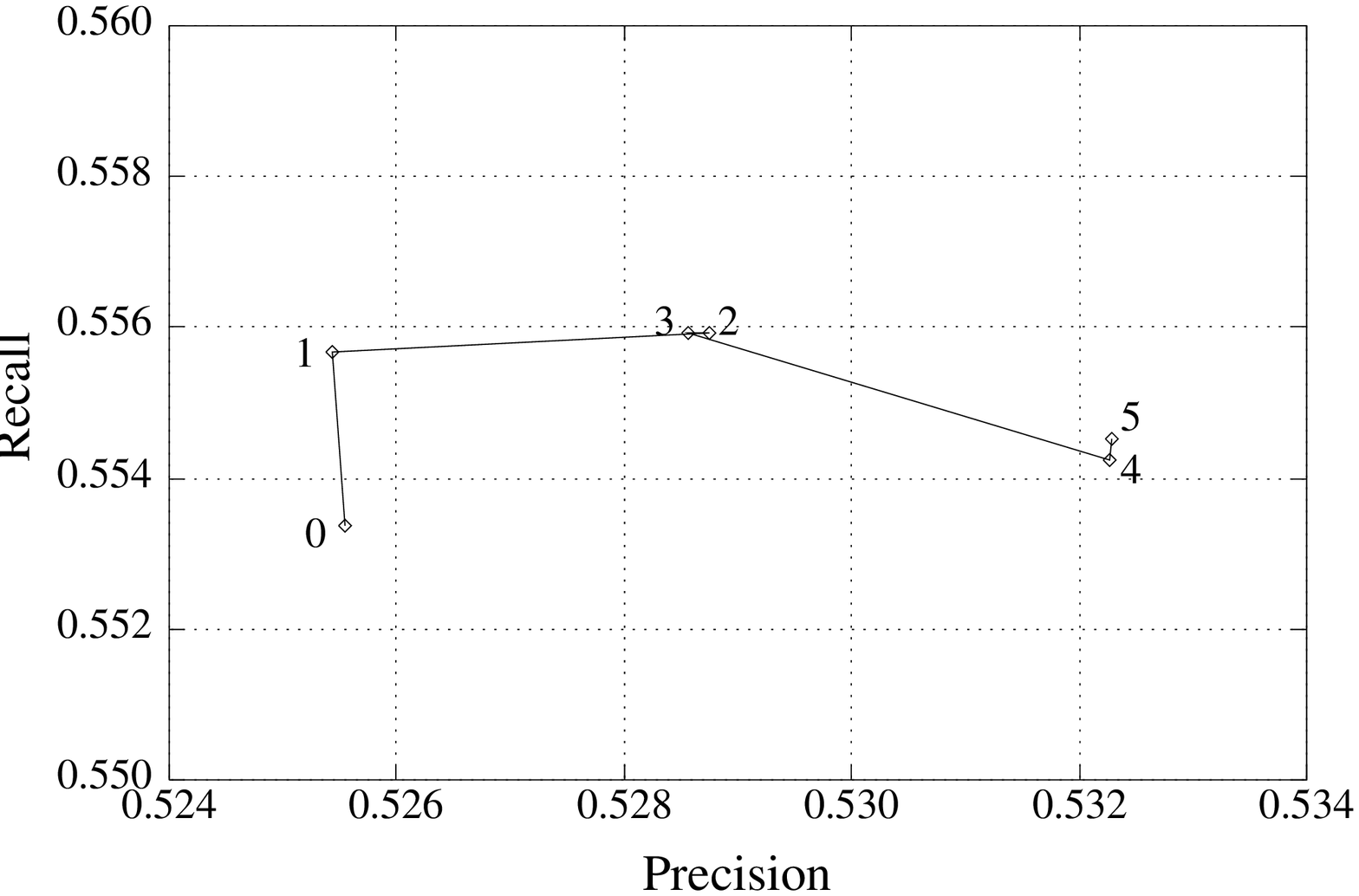,width=3in}}
\caption{{\em French $\rightarrow$ English BiBLE scores for 6
translation models.  Labels 0 to 5 indicate iteration
number.}\label{FEbiblenonulls}}
\end{figure}
\begin{figure}[htb]
\centerline{\psfig{figure=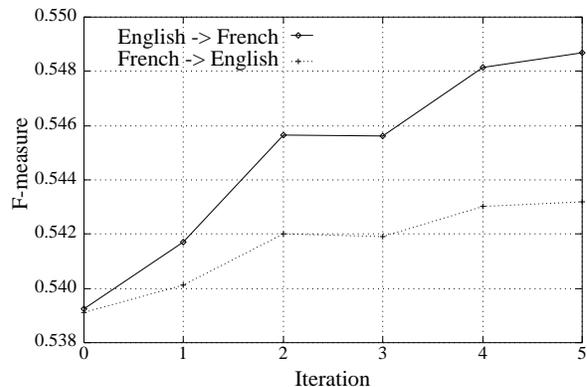,width=3in}}
\caption{{\em F-measures for BiBLE tests on successive
translation models.}\label{Fmeasure.nonulls}}
\end{figure}
The second experiment was based on the simpler objective function $V$
and its associated predictive value function in Equation~\ref{v}.  The
impact of NCC recognition on the bag-of-words translation task
was measured directly, using Bitext-Based \mbox{Lexicon} Evaluation
\citep[BiBLE:][]{mel95}.  BiBLE is a family of evaluation algorithms
for comparing different translation methods objectively and
automatically.  The algorithms are based on the observation that if
translation method $A$ is better than translation method $B$, and each
method produces a translation from one half of a held-out test bitext,
then the other half of that bitext will be more similar to the
translation produced by $A$ than to the translation produced by $B$.
In the present experiment, the translation method was always
bag-of-words translation, but using different translation models.  The
similarity of two texts was measured in terms of word precision and
word recall in aligned sentence pairs, ignoring word order.

I compared the 6 base translation models induced in 6 iterations
of the algorithm in Section~\ref{alg}.\footnote{The entire algorithm
was only run five times, but Steps~2 and~3 were run a sixth time.}  The
first model is numbered 0, to indicate that it did not recognize any
NCCs.  The 6 translation models were evaluated on the test bitext
$(E,F)$ using the following BiBLE algorithm:
\begin{enumerate}
\item Fuse all word sequences in $E$ that correspond to NCCs
recognized by the translation model.
\item Initialize the counters $a$ and $c$ to zero.
\item Let $b$ be the number of words in $F$. 
\item For each pair of aligned sentences \mbox{$(e,f) \in (E,F)$},
\begin{enumerate}
\item For each word $s$ in $e$, add the most likely translation of $s$
to the trial target sentence $\hat{f}$.  If the most likely
translation is an NCC, then break it up into its components.  If $s$
is not in the translation model (an unknown word), then add $s$ itself
to $\hat{f}$.
\item $a = a + |\hat{f}|$ 
\item For each word in $\hat{f}$, check whether it occurs in $f$.  If
so, increment the counter $c$ and remove the word from
$f$.\footnote{Removing words from $f$ in Step 3(c) is necessary to
ensure that no target word gives credit to more than one source word
translation, and thereby to foil a simple method of cheating: If
matched words in $f$ are not removed, then a trivial translation model
where all source words translate to the most frequent target word
would score surprisingly high!  E.g. a French to English translation
method that outputs ``the the the the \ldots'' would recall more than
6\% of English words.}
\end{enumerate}
\item Precision := $c / a$. Recall := $c / b$.
\end{enumerate}

The BiBLE algorithm compared the 6 models in both directions of
translation. The results are detailed in Figures~\ref{biblenonulls}
and~\ref{FEbiblenonulls}.  Figure~\ref{Fmeasure.nonulls} shows
F-measures that are standard in the information retrieval literature:
\begin{equation}
F = \frac{2 * precision * recall}{precision + recall}
\end{equation}
The absolute recall and precision values in these figures are quite
low, but this is not a reflection of the quality of the translation
models.  Rather, it is an expected outcome of BiBLE evaluation, which
is quite harsh. Many translations are not word for word in real
bitexts and BiBLE does not even give credit for synonyms.  The best
possible performance on this kind of BiBLE evaluation has been
estimated at 62\% precision and 60\% recall \citep{mel95}.

The purpose of BiBLE is internally valid comparison, rather than
externally valid benchmarking.  On a sufficiently large test bitext,
BiBLE can expose the slightest differences in translation quality.
The number of NCCs validated on each iteration was never more than
2.5\% of the vocabulary size.  Thus, the curves in
Figures~\ref{biblenonulls} and~\ref{FEbiblenonulls} have a very small
range, but the trends are clear.

A qualitative assessment of the NCC discovery method can be made by
looking at Table~\ref{sample}.  It contains a random sample of 50 of
the English NCCs accumulated in the first five iterations of the
algorithm in Section~\ref{alg}, using the simpler objective function
$V$.  All of the NCCs in the table are non-compositional with respect
to the objective function $V$.  Many of the NCCs, like ``red tape''
and ``blaze the trail,'' are true idioms.  Some NCCs are incomplete.
E.g. ``flow-'' has not yet been recognized as a non-compositional part
of ``flow-{\em through share},'' and likewise for ``head'' in ``{\em
rear its ugly} head.''  These NCCs would likely be completed if the
algorithm were allowed to run for more iterations.  Some of the other
entries deserve more explanation.

First, ``Della Noce'' is the last name of a Canadian Member of
Parliament.  Every occurrence of this name in the French training text
was tokenized as ``Della noce'' with a lowercase ``n,'' because
``noce'' is a common noun in French meaning ``marriage,'' and the
tokenization algorithm lowercases all capitalized words that are found
in the lexicon.  When this word occurs in the French text without
``Della,'' its English translation is ``marriage,'' but when it occurs
as part of the name, its translation is ``Noce.''  So, the French
bigram ``Della Noce'' is non-compositional with respect to the
objective function $V$.  It was validated as an NCC.  On a subsequent
iteration, the algorithm found that the English bigram ``Della Noce''
was always linked to one French word, the NCC ``Della\_noce,'' so it
decided that the English ``Della Noce'' must also be an NCC.  This is
one of the few non-compositional personal names in the Hansards.

Another interesting entry in the table is the last one.  The
capitalized English words ``Generic'' and ``Association'' are
translated with perfect consistency to ``Generic'' and
``association,'' respectively, in the training text.  The translation
of the middle two words, however, is non-compositional.  When
``Pharmaceutical'' and ``Industry'' occur together, they are rendered
in the French text without translation as ``Pharmaceutical Industry.''
When they occur separately, they are translated into
``pharmaceutique'' and ``industrie.''  Thus, the English bigram
``Pharmaceutical Industry'' is an NCC, but the words that always occur
around it are not part of the NCC.

Similar reasoning applies to ``{\em ship unprocessed} uranium.''  The
bigram $<$ {\em ship, unprocessed} $>$ is an NCC because its
components are translated non-compositionally whenever they co-occur.
However, ``uranium'' is always translated as ``uranium,'' so it is not
a part of the NCC.  This NCC demonstrates that valid NCCs may cross
the boundaries of grammatical constituents.

\section{Related Work}

In their seminal work on statistical machine translation, \citet{ibm}
implicitly accounted for NCCs in the target language by estimating
``fertility'' distributions for words in the source language.  A
source word $s$ with fertility $n$ could generate a sequence of $n$
target words, if each word in the sequence was also in the
translational distribution of $s$ and the target language model
assigned a sufficiently high probability to the sequence.  However,
\citeauthor{ibm}'s models do not account for NCCs in the source
language.  Recognition of source-language NCCs would certainly improve
the performance of their models, but \citeauthor{ibm} warn that
\begin{quote}
\ldots one must be discriminating in choosing multi-word cepts.  The
caution that we have displayed thus far in limiting ourselves to cepts
with fewer than two words was motivated primarily by our respect for
the featureless desert that multi-word cepts offer a priori. \citep{ibm}
\end{quote}
The heuristics in Section~\ref{est} are designed specifically to find
the interesting features in that featureless desert.  Furthermore,
translational equivalence relations involving explicit representations
of target-language NCCs are more useful than fertility distributions
for applications that do translation by table lookup.

Many authors \citep[e.g.][]{daille,smadja} define ``collocations'' in
terms of monolingual frequency and part-of-speech patterns.  Markedly
high frequency is a necessary property of NCCs, because otherwise they
would fall out of use.  However, at least for translation-related
applications, it is not a sufficient property.  Non-compositional
translation cannot be detected reliably without looking at
translational distributions.  The deficiency of criteria that ignore
translational distributions is illustrated by their propensity to
validate most personal names as ``collocations.''  At least among West
European languages, translations of the vast majority of personal
names are perfectly compositional.

Several authors have used mutual information and similar statistics as
an objective function for word clustering
\citep{dagan,ngramclus,fernando,langmod}, for automatic determination
of phonemic baseforms \citep{lucassen}, and for language modeling for
speech recognition \citep{ries}.  Although the applications considered
in this paper are different, the strategy is similar: search a space
of data models for the one with maximum predictive power.
\citet{langmod} also employ parallel texts and independence
assumptions that are similar to those described in Section~\ref{est}.
Like \citet{ngramclus}, they report a modest improvement in model
perplexity and encouraging qualitative results.  Unfortunately, their
estimation method cannot propose more than ten or so word-pair
clusters before the translation model must be re-estimated.  Also, the
particular clustering method that they hoped to improve using parallel
data is not very robust for low frequencies.  So, like
\citeauthor{smadja}, they were forced to ignore all words that occur
less than five times.  If appropriate objective functions and
predictive value functions can be found for these other tasks, then
the method in this paper might be applied to them.

There has been some research into matching \mbox{{\em compositional}}
phrases across bitexts.  For example, \citet{kupiec} presented a
method for finding translations of whole noun phrases.  \citet{wu95}
showed how to use an existing translation lexicon to populate a
database of ``phrasal correspondences'' for use in example-based MT.
These compositional translation patterns enable more sophisticated
approaches to MT.  However, they are only useful if they can be
discovered reliably and efficiently.  Their time may come when we have
a better understanding of how to model the human translation process.

\newpage
\section{Conclusion}

It is well known that two languages are more informative than one
\citep{langinf}.  I have argued that texts in two languages are not
only preferable but necessary for discovery of non-compositional
compounds for translation-related applications.  Given a method for
constructing statistical translation models, NCCs can be discovered by
maximizing the models' information-theoretic predictive value over
parallel data sets.  This paper presented an efficient algorithm for
such ontological discovery.  Proper recognition of NCCs resulted in
improved performance on a simple MT task.

Lists of NCCs derived from parallel data may be useful for NLP
applications that do not involve parallel data.  Translation-oriented
NCC lists can be used directly in applications that have a human in
the loop, such as computer-assisted lexicography, computer-assisted
language learning, and corpus linguistics.  To the extent that
translation-oriented definitions of compositionality overlap with
other definitions, NCC lists derived from parallel data may benefit
other applications where NCCs play a role, such as information
retrieval \citep{ir} and language modeling for speech recognition
\citep{ries}.  To the extent that different applications have
different objective functions, optimizing these functions can benefit
from an understanding of how they differ.  The present work was a step
towards such understanding, because ``an explication of a monolingual
idiom might best be given after bilingual idioms have been properly
understood'' \citep[p. 48]{bh}.

The NCC discovery method makes few assumptions about the data sets
from which the statistical translation models are induced.  As
demonstrated in Section~\ref{results}, the method can find NCCs in
English letter strings that are aligned with their phonetic
representations.  We hope to use this method to discover NCCs in other
kinds of parallel data.  A natural next target is bitexts involving
Asian languages.  Perhaps the method presented here, combined with an
appropriate translation model, can make some progress on the word
identification problem for languages like Chinese and Japanese.

\newpage
\begin{table*}[H]
\centering
\begin{tabular}{|r|l|l|}
\hline
Count & NCC (in italics) in typical context & non-compositional
translation in French text \\ 
\hline\hline
786	& {\em could have} &  pourrait \\
183	& flow-{\em through shares} & actions accr\'{e}ditives  \\
79	& {\em I repeat} &  je tiens \`{a} dire \\
63	& the case I {\em just mentioned} & le cas que je viens de mentionner  \\
36	& {\em tax base} &  assiette fiscale \\
34	& {\em single parent} family & famille monoparentale  \\
24	& {\em perform $<GAP>$ duty} & assumer \ldots fonction \\
23	& {\em red tape} & la paperasserie \\
17	& {\em middle of the night} & en pleine nuit  \\
17	& {\em Della Noce} &  Della noce (see text for explanation) \\
16	& {\em heating oil} & mazout  \\
14	& {\em proceeds of crime} & les produits tir\'{e}s du crime  \\
11	& {\em rat pack} & meute  \\
10	& {\em urban dwellers} & citadins  \\
10	& nuclear {\em generating station} & centrale nucl\'{e}aire\\
10	& Air {\em India disaster} & \'{e}crasement de l'avion indien \\
9	& {\em Ottawa River} & Outaouais  \\
8	& I {\em dare hope} &  j'ose croire \\
8	& {\em Ottawa Valley} & vall\'{e}e de l'Outaouais  \\
7	& {\em plea bargaining} & marchandage  \\
7	& {\em manifestly unfounded} claims & avoir revendiqu\'{e} \'{a} tort le statut  \\
7	& {\em machine gun} &  mitrailleuse \\
7	& a group called {\em Rural Dignity} & une groupe appel\'{e} Rural Dignity  \\
6	& a {\em slight bit} &  la moindre \\
6	& {\em cry for help} & appel au secour  \\
5	& {\em video tape} & vid\'{e}o  \\
5	& {\em sow the seed} & semer  \\
5	& {\em arrange a meeting} & organiser un entretien  \\
4	& {\em shot-gun wedding} & mariage forc\'{e}  \\
4	& we {\em lag behind} & nous tra\^{i}nions de la patte  \\
4	& {\em Great West} Life Company & Great West Life Company  \\
4	& Canadian Forces {\em Base and cease} negotiations &  mettre fin et interrompre le n\'{e}gociation  \\
3	& {\em severe sentence} & s\'{e}v\`{e}re sanction  \\
3	& {\em rear its ugly} head & manifest\'{e}  \\
3	& {\em inability to deal} effectively with  & ne sait pas traiter de mani\`{e}re efficace avec  \\
3	& {\em en masse} & en bloc  \\
3	& {\em create a disturbance} & suscite de perturbation  \\
3	& {\em blaze the trail} & ouvre la voie   \\
2	& {\em wrongful conviction} & erreur judiciaire  \\
2	& {\em weak sister} & parent pauvre  \\
2	& of both the {\em users and providers} of transportation & des utilisateurs et des transporteurs  \\
2	& {\em understand the motivation} & saisir le motif  \\
2	& {\em swimming pool} & piscine  \\
2	& {\em ship unprocessed} uranium & exp\'{e}dier de l'uranium non raffin\'{e} \\
2	& by {\em reason of insanity} & pour cause d'ali\'{e}nation mentale  \\
2	& l'agence de Presse {\em libre du} Qu\'{e}bec & l'agence de Presse libre du Qu\'{e}bec  \\
2	& do {\em cold weather} research &  \'{e}tudier l'effet du froid \\
2	& the {\em bread basket} of the nation  & le grenier du Canada \\
2	& turn back the {\em boatload of European} Jews & renvoyer tout ces juifs europ\'{e}ens  \\
2	& Generic {\em Pharmaceutical Industry} Association &  Generic Pharmaceutical Industry Association \\
\hline
\end{tabular}
\caption{{\em Random sample of 50 of the English NCCs validated in
the first five iterations of the NCC discovery algorithm, using the
objective function V.  ``Count'' is the number of times the NCC occurs in
the training text.  All the NCCs are non-compositional with respect to
the objective function $V$.}
\label{sample}}
\end{table*}

\end{document}